\documentclass[sigconf]{acmart}
\AtBeginDocument{%
  }

\usepackage{listings}
\usepackage{multirow}
\usepackage[utf8]{inputenc}
\usepackage{enumitem}

\copyrightyear{2025}
\acmYear{2025}
\setcopyright{acmlicensed}\acmConference[GeoGenAgent '25]{The 1st ACM SIGSPATIAL International Workshop on Generative and Agentic AI for Multi-Modality Space-Time Intelligence}{November 3--6, 2025}{Minneapolis, MN, USA}
\acmBooktitle{The 1st ACM SIGSPATIAL International Workshop on Generative and Agentic AI for Multi-Modality Space-Time Intelligence (GeoGenAgent '25), November 3--6, 2025, Minneapolis, MN, USA}
\acmDOI{10.1145/3764915.3770721}
\acmISBN{/2025/11}

\usepackage{comment}
\usepackage{verbatim}
\usepackage{tcolorbox}
\usepackage{underscore}  % This package handles underscores in text mode
\usepackage{xcolor}
\newcommand{\tool}[1]{\texttt{#1}}
\newcommand{\user}[1]{\textit{``\textcolor{blue!30!black}{#1}}''}
\begin{document}

\title{GeoBenchX: Benchmarking LLMs in Agent Solving Multistep Geospatial Tasks}

\author{Varvara Krechetova}
\affiliation{%
  \institution{World Bank}
  \city{New York}
  \country{USA}
}

\author{Denis Kochedykov}
\affiliation{%
  \institution{JP Morgan Chase \& Co}
  \city{New York}
  \country{USA}
}

\renewcommand{\shortauthors}{Krechetova, Kochedykov}

\begin{abstract}
This paper establishes a benchmark for evaluating tool-calling capabilities of large language models (LLMs) on multi-step geospatial tasks relevant to commercial GIS practitioners. We assess eight commercial LLMs (Claude Sonnet 3.5 and 4, Claude Haiku 3.5, Gemini 2.0 Flash, Gemini 2.5 Pro Preview, GPT-4o, GPT-4.1 and o4-mini) using a simple tool-calling agent equipped with 23 geospatial functions. Our benchmark comprises tasks in four categories of increasing complexity, with both solvable and intentionally unsolvable tasks to test rejection accuracy. We develop a  LLM-as-Judge evaluation framework to compare agent solutions against reference solutions. Results show o4-mini and Claude 3.5 Sonnet achieve the best overall performance, OpenAI's GPT-4.1, GPT-4o and Google's Gemini 2.5 Pro Preview do not fall far behind, but the last two are more efficient in identifying unsolvable tasks. Claude Sonnet 4, due its preference to provide any solution rather than reject a task, proved to be less accurate. We observe significant differences in token usage, with Anthropic models consuming more tokens than competitors. Common errors include misunderstanding geometrical relationships, relying on outdated knowledge, and inefficient data manipulation. 
The resulting benchmark set, evaluation framework, and data generation pipeline are released as open-source resources \footnote{Available at https://github.com/Solirinai/GeoBenchX}, providing one more standardized method for the ongoing evaluation of LLMs for GeoAI.
\end{abstract}



% %%


% \received{20 September 2025}
% % \received[revised]{2025}
% \received[accepted]{29 September 2025}

%%
\maketitle

\section{Introduction}
In this paper, we create a benchmark set and describe a simple LLM-as-Judge based methodology to compare the tool-calling performance of several leading commercial large language models (LLMs) with function-calling abilities for solving a range of multi-step geospatial tasks. Geospatial tasks are distinct in that they require LLMs to have good factual geographical knowledge, understanding how to work with spatially defined data, what are coordinate reference systems, and various types of spatial data. We used a simple tool-calling agent to specifically assess the LLMs capabilities and have tested several models: Anthropic's Claude Sonnet 3.5, 4, Claude Haiku 3.5; Google's Gemini 2.0 Flash, Gemini Pro 2.5 Preview; OpenAI's GPT-4o, GPT-4.1, o3-mini and o4-mini. 

Our focus is on the commercial GIS practitioner requirements, which influences the choice of tasks, models and evaluation framework. Unlike other adjacent benchmarks, we focus on tasks needed for socio-economic analysis and mapping for reports, policy papers, news articles,  and the like: tasking agent to make various types of maps, to find geographical objects by some conditions, to count objects within certain geographic boundaries. For these tasks, practitioners would use LLM equipped with GIS tools in a chat environment. Also, importantly for practitioners, we test the ability of LLMs to correctly recognize when the task cannot be solved with given tools and inform the user to reduce the risk of hallucinated outputs. 
Next, we selected the most current advanced commercial LLMs used in the industry setting for benchmarking. 
Lastly, the evaluation method we use allows for automation of the ongoing evaluation which is increasingly important with frequent models updates in commercial settings.

The paper is structured as follows. First, we describe the task-solving agent and the set of tools. Second, we describe how the benchmark set of tasks was created and its composition. Third, we detail the approach to evaluation of the results of the task-solver agent using a panel of LLM-as-Judge evaluators and testing the evaluator performance itself. Finally, we compare the performance metrics of the solver agent based on different LLMs and the cost efficiency in terms of token usage.

\section{Related works}
Research in the field of GeoAI, a part of spatial data science at the intersection with Artificial Intelligence (AI) \cite{Janowicz02042020}, got a boost with recent advances in generative AI and the deployment of accessible LLMs. GeoAI researchers actively explore the capabilities of the available models in relation to geospatial data science, design their applications, and methods to improve their performance. The avenues explored include analysis of the ability of LLM (GPT-4) for spatial reasoning in travel networks, distance and elevation estimations, route and supply chain analysis \cite{roberts2023gpt4geo}; integration of geography knowledge in pre-trained language models \cite{RAMRAKHIYANI2025103892}, creating autonomous LLM-based GIS applications (LLM-Geo) \cite{Li08122023}, GIS Copilot for GIS software \cite{gis-copilot}, integrating LLM and cartography tools into MapGPT \cite{zhangMapGPT}, testing LLMs in styling maps using reference visuals \cite{wang2025cartoagent}, combining several LLMs to increase the efficiency of spatial data processing in the multi-agent framework for automated processing of vector data in shapefile format \cite{lin2024shapefilegpt}, benefits of generative AI agents in smart mobility\cite{AI4ITS-xu2024}, fine-tuning LLM to process natural language queries into executable code\cite{ijgi13100348}.
There is a benchmarking of LLM performance in spatial tasks in progress, with analysis of LLMs' geospatial knowledge\cite{LLM-geoknowledge, Renshaw04032025}, of GIS coding abilities\cite{Gramacki-2024}, the recent study testing models on several groups of tasks that require geographic, GIS, and spatial knowledge, including changes in their performance in response to changes in prompts \cite{zu2025-llm-eval}. 

\section{Experiment setup}
\subsection{Agent}
The experiment setup is based on the pre-built Langgraph ReAct agent \cite{langgraphReact2024} equipped to call one of the selected LLMs with a list of tools (functions) to solve tasks. We intentionally choose a simple agent architecture to measure the contribution of the LLMs themselves into the final performance, not compensated and leveled by complex agent architectures. Typical practical requirements are that the variability in outputs is undesirable, but consistency in responses is critical, so the models were tested at temperature 0, which reduces variability and increases consistency of outputs \cite{openai_api_temperature, google_vertex_ai_function_calling}.

The agent processed a given user task without Human-In-The-Loop involvement. Each task normally requires series of tool calls to solve and the agent does not always find the shortest solutions, so we allowed up to 25 iterations of ``LLM suggests a tool call --> LLM receives the tool response''. This number was selected after initial trials of the agent that showed that fewer iterations were not enough for some complex tasks, while allowing more steps did not increase the success rates (accuracy).

The agent is supplied with a simple system prompt that sets the persona, briefly describes the setting, and the rules for solving the tasks. The agent is explicitly instructed not to ask the user for clarifications and not to provide code to solve the task, but to reply with the resulting map, names of the datasets used, and explanations. The set of rules contains instructions on when to reject the task and when to leave the solution empty (meaning that no tools are needed to answer user's question), when to consider the data unavailable (if the available data do not fully cover the geography asked in the task), and how to select columns for mapping data if the user did not specify (implied conditions frequent in production user questions).

The text of the task is supplied as the next message for the LLM's Chat API as the user prompt. 
\subsection{Agent tools}
\label{subsec-tools}
To enable the agent to solve the tasks, it is provided with 23 tools to implement the necessary data manipulations, calculations, and visualizations or to report a task unsolvable. Tools can read tabular data, geospatial data in vector or raster formats, perform operations on dataframes and geodataframes (merge, filter by categorical or numerical values), their columns, perform spatial operations on geodataframes and rasters (spatial joins, filtering by locations, creating buffers, generate contour lines from raster), visualize geodataframes, generate choropleth maps and heatmaps. The tools leverage GIS-specific libraries like GeoPandas \cite{joris-van-den-bossche-2024-12625316}, Rasterio \cite{gillies-2019}, GDAL \cite{rouault-2025-14639689}, Shapely \cite{gillies-2025-14776272}, and common data analysis libraries like Matplotlib, Plotly, Pandas, and Numpy. To distinguish the situations in which the agent stops solving the task because it thinks that the task could not be solved with the datasets / tools provided, the agent was equipped with an additional tool \tool{reject_task} that simply returned the message that the task is unsolvable.

The outputs of the tool functions were formatted as natural part of the tool response message. When a tool result included a dataframe or geodataframe, information about its length, columns, data types, and non-empty cells was included in the tool response message. Where dataframes or geodataframes needed to be passed between tools, they were stored in the data-store dictionary accessible to all tools. 

For clarity, hereafter we refer to the set of tasks used to evaluate the agent as ``benchmark set'' and the statistical and geodata used by the tools to solve the tasks as ``datasets''.

To avoid variability of benchmarking due to possible data API connection issues, data API changes, and data changes behind the scenes, we opt to store the snapshots of datasets in the local experimentation environment rather than fetching data from online APIs every time. 

% Snapshots of datasets were organized as dictionaries with short descriptive titles as keys and filenames as values, one dictionary per type of data. When the tool designed to load the respective data was called, LLM had to choose the tool argument, the dataset to load, from the list of the keys of the appropriate dictionary. 

For each task, the conversation history between LLM and tools along with any plotted outputs is saved as html file making the final LLM response, maps and intermediate outputs available for debugging.
\subsection{Datasets}
The datasets were used as-is after downloading from online APIs, except for two modifications made to ease development: country names in the vector geodataset were harmonized with statistical datasets for easier visual identification of filtering issues, and the NoData value in snow cover rasters was unified. These changes were not required for the benchmarking itself.

The data included:

{\bfseries Statistical data:} panel data in csv format (18 datasets): data on economic and population indicators, data on CO and greenhouse gas emissions, tuberculosis cases at subnational levels.

{\bfseries Geometry data:} vector data in shapefile format (21 datasets): polygon geometries (national or subnational units, lakes), points geometries (wildfire locations, urban places, Amtrak stations, earthquake occurrences, seaports, mineral extraction facilities, power stations at global or national level), line geometries (rivers, railways continent or national extent).

{\bfseries Raster data:} (11 rasters): accumulated snow cover, flood extents, population data at national level.

\section{Benchmark set}

We created a set of 200+ tasks, grouped by type and complexity. For estimating a binary match rate at the level of match we see in results section (\ref{sec-eval-res}) -- this sample size corresponds approximately to 5\% margin of error with 90\% confidence according to Wald method. Each group included tasks that could be solved with the tools and datasets provided (solvable tasks), tasks that looked similar but could not be solved with the tools or datasets provided (unsolvable tasks). For debugging purposes, we also added a few tasks that did not require use of the provided tools or datasets to solve. 
\subsection{Groups}
\label{sec-groups}
Depending on the complexity of data processing or geoprocessing, the task groups are as follows:

{\bfseries Merge - visualize} (36 tasks): require to join tabular numeric data (panel data for one or multiple indicators, e.g. countries-by-years) with a geographic geometry data and make a choropleth or bivariate map. Example: \user{Map the relationship between GDP per capita and electric power consumption per capita globally}. This is the most straightforward group, the tasks mostly require two-three tool calls. Only 6 of these tasks were not solvable with given tools/data. Due to simplicity of the tasks, there is limited variation in their content and solutions and a smaller number provided enough coverage.

{\bfseries Process - merge - visualize} (56 tasks): require some processing of either tabular or geometry data (filtering or operations on columns) and then creating a choropleth or bivariate map. Example: \user{Compare rural population percentages in Western Asian countries}. The tasks in this group are more complicated, require the use of knowledge of the world geography and the use of a wider variety of tools. About half of the tasks, 24, are not solvable with given tools/data.

{\bfseries Spatial operations} (53 tasks): require spatial joins, creating buffers around selected geometries, checking overlaps between rasters, performing calculations on rasters, finding values from raster using geometry data or filtering geometries based on raster values, calculating distances. They might require a numerical answer, and the creation of actual maps might be optional. This group is more challenging than the previous two, as it requires the agent to work with different types of datasets in one task, navigate the full set of available tools, and maintain the overall logic of the solution through all its steps.  Example: \user{How many people live within 1 km from a railway in Bangladesh?}. In this group, 22 of the tasks were not solvable.

{\bfseries Heatmaps, contour lines} (54 tasks): required performing spatial operations (selection, overlaps, etc.), understanding the contents of rasters, and making a heatmap or generating contour lines as one of the steps. Example: \user{Create a heatmap of earthquake occurrences in regions with high population growth}. This was the most challenging set, as the agent had to figure out if the heatmaps and contour lines tools were actually able to solve the task and to make correct decisions on a complex set of arguments. In this group, 27 tasks were not solvable.

\subsection{Tasks creation}
For each group, the seed tasks were sourced from industry professionals and publications. Then more tasks were generated by using the Claude 3.5 Sonnet model based on: the seed tasks, description of the tools, and the lists of available datasets. Unsolvable and solvable tasks were generated separately. The generated tasks were then manually reviewed to keep only high-quality tasks in the set.

\subsection{Tasks annotation}
\label{sec-annotate}
To allow the evaluation of LLM-generated solutions, each task was annotated with the ground truth, reference solution, formatted as snippets of Python code (see more in Section \ref{eval-approach}). 
To create the annotation, we have started by generating draft reference solutions by the agent using GPT‑4o or Claude 3.5 Sonnet models. We then manually reviewed and rewrote every solution. Each of the 200+ tasks was annotated with one or multiple ground-truth reference solutions. 

Reference solution could be 
\begin{enumerate}%[leftmargin=*, labelindent=0pt]
\label{ref-sol}
    \item Sequence of tool-calls with arguments; similar to a python code snippet 
    \item Empty sequence \lstinline|[]| of tool-calls meaning that the task can be solved from the LLM's world knowledge
    \item A single special tool call \lstinline|reject_task| meaning that the task should be rejected as it cannot be solved with given tools/data
\end{enumerate}

A non-empty reference solution looked as follows for the evaluator:
\begin{lstlisting}[language=Python,  basicstyle=\ttfamily\small,identifierstyle=\color{blue}, commentstyle=\color{gray}]
#overall reference solution comment
toolcall1(arg11=..., arg12=..., ...) #call1 comment
toolcall2(arg21=..., arg12=..., ...) #call2 comment
...
\end{lstlisting} 

One task sometimes allows multiple solutions because tasks in practice are often ambiguous and multiple analytical approaches are possible for them based on the provided data and tools. For example, task \user{Compare the freshwater withdrawal between African countries with and without significant railway networks} can be solved:
\begin{itemize}
    \item by rejecting because it's not specific enough (user needs to clarify what's ``significant'' railway networks);
    \item by making separate maps of railway networks and water withdrawals for African countries;
    \item by combining the data and mapping bivariate map to show the relation;
    \item by implying some threshold and computing statistical metrics for 2 groups of countries.
\end{itemize}

We also used comments as a way to record some information important for the evaluator to make a judgment about whether the candidate solution and the reference solution are semantically equivalent.
For example, providing alternative arguments to the tool call, specifying variations in arguments or marking optional steps.

Approximately 74\% of the solvable tasks have reference solutions consisting of 4-6 steps (tool-calls) and ~28\% consist of more steps (up to 14 steps). The numbers do not total to 100\% as a task can have multiple reference solutions.
\section{Evaluation approach}
\label{eval-approach}
\subsection{Matching-based evaluator}
To leverage the LLMs ability to comprehend code \cite{nikiema2025codebarrierllmsactually, haroon2025accuratelylargelanguagemodels, ma2024lmsunderstandingcodesyntax}, we represented both reference and candidate solutions as snippets of Python code as in \ref{ref-sol}.
In code evaluation, the two widely used approaches are: execution-based and matching-based, both used for the evaluation of LLM-generated code \cite{10403378, 10189234}. Since the outputs of agent on geospatial analytical tasks contain multiple images/maps, numbers, a narrative, and sometimes all of it as in the section \ref{sec-annotate} example, reliably comparing them via execution-based approach would be challenging and annotation of tasks with such ground truth is rarely practical. We opt for the matching-based approach for practicality and efficiency. 
We used reference-based LLM-as-Judge with multiple LLM judges to compare the solutions generated by the agent (``candidate solutions'') and reference solutions for the same task. Compared to human evaluation, the use of LLMs to compare the generated solution and reference is cost- and time-efficient \cite{chianglee2023large}; if multiple rounds of evaluation are needed due to changes in tasks, their number, or evaluated LLMs, it offers reproducibility \cite{chianglee2023large} and scalability \cite{judgingLLMjudge}.

We asked LLM judge to compare semantic equivalence of the solution; this took care of situations where reference and candidate solutions had different order of steps, different arguments for tool-calls or even took a different path to the same analytical outcome.

We also compared the lengths of the reference and candidate solutions, since the agent often takes some steps to figure out a path to answering the user question, and such extra steps might be costly (see \ref{subsec-tokens}).
Looking into solution code also provides interesting insights into how LLM solves the tasks and allows agent tuning (which we did not do for this benchmarking).

For evaluation of the solutions we used a panel of judges: Claude Sonnet 3.5, GPT-4.1 and Gemini 2.5 Pro Preview. The evaluator prompt included: persona setting, description of all tools, scoring taxonomy, instructions on how evaluate solutions, examples of evaluated solutions, corresponding scores and reasoning behind each score. 
The evaluator was instructed to analyze the candidate solution for the task and produce 
(1) reasoning for matching any of the reference solutions for the task and the candidate solution and (2) the matching score  ('2': match, '0': not a match, '1': partial match).

\subsection{Evaluating the evaluator}
We evaluated and tuned the evaluator itself on the set of 50 tasks (about 25\% of the benchmark set) randomly sampled with stratification from all tasks, respecting the task groups. 
We manually scored candidate solutions for these tasks, providing a match score and reasoning for it. 
We then ran the evaluator on the same tasks candidate solutions and we compared human annotator match scores and evaluator produced match scores.

\begin{figure}
    \centering
    \includegraphics[width=\linewidth]{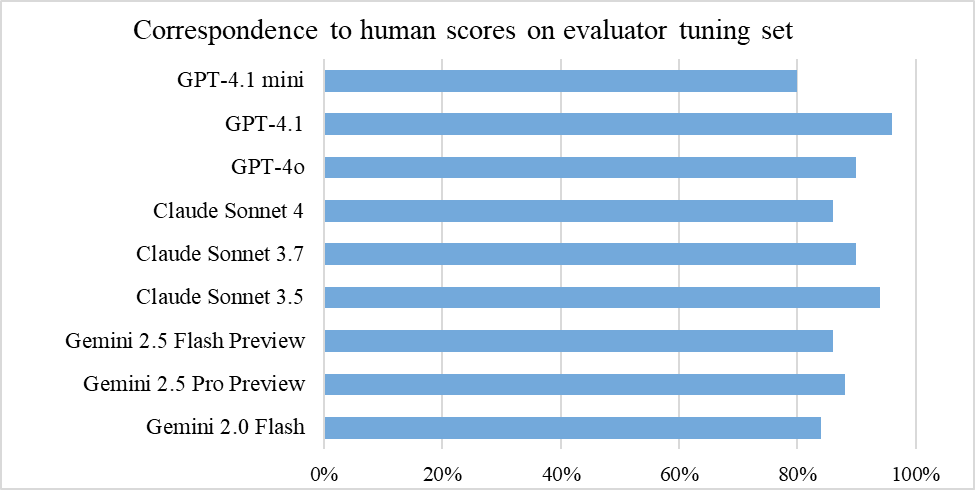}
        \caption{Testing of LLMs as judges}
    \label{fig:testing-judges}
    \Description{Bar chart showing correspondence in percent}
\end{figure}

Nine models were tested as judges (Gemini 2.0 Flash, Gemini 2.5 Pro Preview, Gemini 2.5 Flash Preview; Claude Sonnet 3.5, 3.7, 4; GPT-4o, GPT-4.1, GPT-4.1 mini) and from each family of models the one with the results closest to human annotator scores was selected for the panel: GPT-4.1 with 96\% of scores matching the manual scores, Claude Sonnet 3.5 (94\%) and Gemini 2.5 Pro Preview (88\%) (see Figure~\ref{fig:testing-judges}). The top models of each family were selected rather than simply the top three models to reduce variance in the evaluation results due to differences in training and tuning of the models by different providers.

In the evaluator, we use all the 3 above LLMs as judges and average their match frequency estimates. 

\section{LLMs evaluation results}
\label{sec-eval-res}
\subsection{Experiment procedure}
Nine models were tested in the solver agent: gemini-2.0-flash-001, gemini-2.5-pro-preview-05-06, claude-3-5-haiku-20241022, claude-3-5-sonnet-20241022, claude-sonnet-4-20250514, gpt-4o-2024-08-06, gpt-4.1-2025-04-14, o3-mini-2025-01-31, o4-mini-2025-04-16.
Each LLM was tested on the same benchmark set of tasks, at temperature 0 (except o3-mini, o4-mini which do not support the temperature parameter). 

The frequency of each of the match scores was calculated: 
\begin{enumerate}
    \item separately by each group of tasks as per \ref{sec-groups} and
    \item for solvable tasks (123 tasks) and unsolvable tasks (79 tasks) across all groups.
\end{enumerate}
% To mark unsolvable tasks (tasks with only one reference solution which is to reject the task), the agent was provided with a \tool{reject_task} tool, that it could call either upfront or after some steps if it reached the conclusion that the task cannot be solved with available datasets and tools.

To benchmark how efficient an LLM is at solving a task compared to the reference solution, we also calculated the mean difference between the length (number of steps) of the reference solution for the task (or average length of reference solutions if multiple) and the candidate solution. We express this difference relative to the length of candidate solution (so 0.5 means 50\% more steps in the candidate solution).

The o3-mini model generated empty solutions (no tool calls, just thinking) for 103 tasks and rejected 75 tasks, and while the accuracy in identification of unsolvable tasks thus was 0.75, the model was excluded from further analysis due to obvious inefficiency in using tools to solve tasks.
\subsection{Performance on solvable and unsolvable tasks}
\begin{figure*}
    \centering
    \includegraphics[width=1\linewidth]{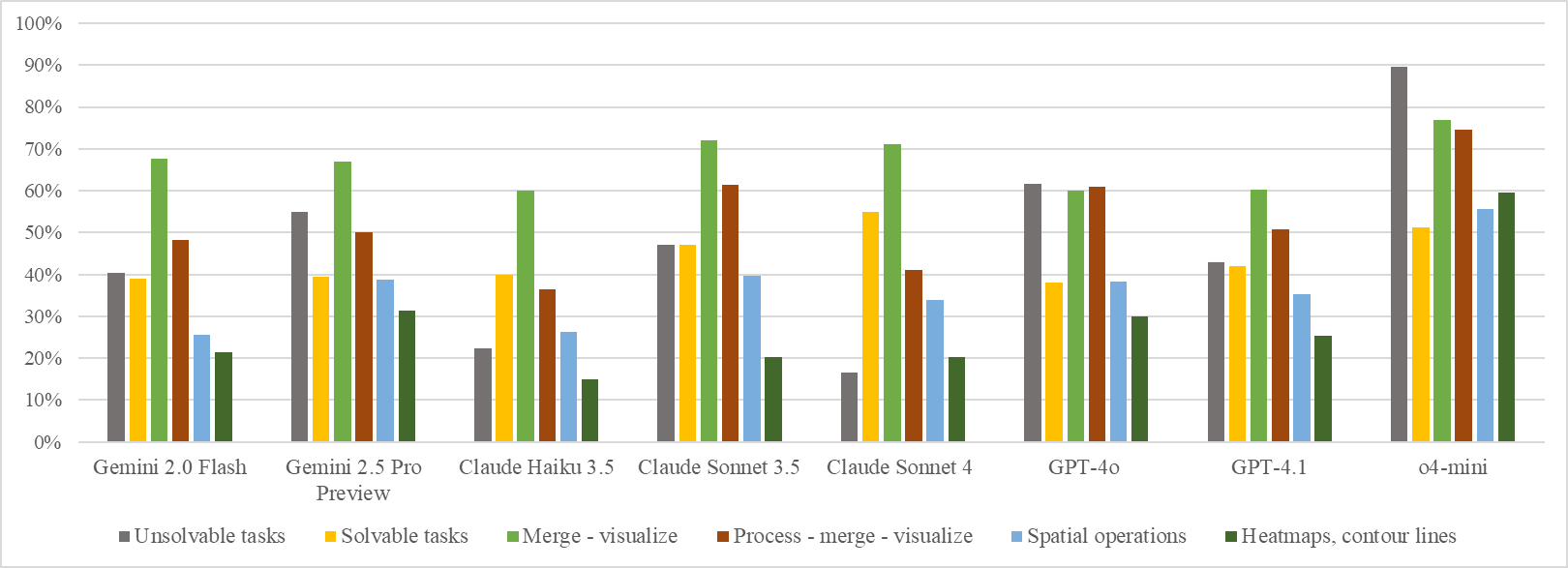}
    \caption{Success rate (share of matches with reference solutions) across LLMs and task types}
    \Description{Column diagram}
    \label{fig:summary-matches}
\end{figure*}

The first observation is that some models demonstrate a trade-off between the ability to identify and reject unsolvable tasks and the ability to solve tasks (Table~\ref{tab:unsolve}, Table~\ref{tab:solve}, Table~\ref{tab:overall}, Figure~\ref{fig:summary-matches}). Claude Sonnet 4 and Haiku 3.5 perform better on solvable tasks but prioritize giving any answer over rejecting tasks, with Claude Sonnet 4 being a prominent case with a success rate on solvable tasks more than three times higher than on unsolvable ones. Gemini 2.5 Pro Preview is somewhat better at detecting unsolvable tasks than solving solvable ones. GPT-4o and o4-mini are substantially better in detecting unsolvable tasks than in solving feasible ones. In each family of models, there is one that demonstrated similar accuracy on both types of tasks in the experiment: Claude Sonnet 3.5, Gemini 2.0 Flash and GPT-4.1.

We can also see, that for unsolvable tasks, the candidate solutions are much longer indicating LLMs trying to find a solution and eventually giving up. For solvable tasks, candidate solutions are about the same length as reference ones, with the exception of Claude Sonnet 4 which tends to produce longer solutions than needed.  

Overall, in solvable tasks, Claude Sonnet 4 turned out to be the best model, followed by o4-mini, Claude Sonnet 3.5, GPT-4.1, and the rest of the models were very close in their success rates (Table~\ref{tab:solve}). However, the highest success rate is only slightly more than half (0.55 and 0.51), and most of the tested models are successful in less than half of the cases (note again that we specifically chose a trivial agent architecture to benchmark the ability of LLMs themselves rather than a particular agentic architecture). Some models have a relatively high proportion of solutions that partially match ground truth solutions (0.16-0.22 for Claude Haiku 3.5, GPT-4.1, Claude Sonnet 4, and Gemini 2.5 Pro Preview), meaning these solutions are very close to correct and models success rates might be increased with minor changes to the agent and tools. 

\begin{table}
  \caption{Success rate in identifying the unsolvable tasks}
  (79 tasks from total of 202)
  \label{tab:unsolve}
  \begin{tabular}{lcccc} 
    \toprule
    Model&No&Partial&Match&$\Delta$ in \\
     &match&match& &solution\\
    &&&&length\\  
    \midrule
    Gemini 2.0 Flash&0.60&0.00&0.40&2.3\\
    Gemini 2.5 Pro Preview&0.45&0.00&0.55&2.0\\        
    Claude Haiku 3.5&0.78&0.00&0.22&4.2\\
    Claude Sonnet 3.5&0.53&0.00&0.47&2.5\\  
    Claude Sonnet 4&0.83&0.00&0.17&7.6\\   
    GPT-4o&0.38&0.00&0.62&2.5\\
    GPT-4.1&0.57&0.00&0.43&4.5\\
    o3-mini&0.25&0.00&0.75&0.0\\  
    o4-mini&0.10&0.00&0.90&0.5\\ 
  \bottomrule
\end{tabular}
\end{table}

\begin{table}
  \caption{Success rate in solvable tasks}
  \label{tab:solve}
   (123 tasks from total of 202)
  \begin{tabular}{lcccc} 
    \toprule
    Model&No&Partial&Match&$\Delta$ in \\
     &match&match& &solution\\
    &&&&length\\  
    \midrule
    Gemini 2.0 Flash&0.52&0.08&0.39&0.0\\
    Gemini 2.5 Pro Preview&0.44&0.16&0.39&0.0\\        
    Claude Haiku 3.5&0.38&0.22&0.40&0.2\\
    Claude Sonnet 3.5&0.41&0.12&0.47&-0.1\\  
    Claude Sonnet 4&0.27&0.18&0.55&0.7\\   
    GPT-4o&0.51&0.12&0.38&0.1\\
    GPT-4.1&0.38&0.20&0.42&0.4\\
    o4-mini&0.38&0.11&0.51&0.1\\ 
  \bottomrule
\end{tabular}
\end{table}
\subsection{Performance by task groups}
We then examine LLMs performance by task groups without distinguishing between solvable and unsolvable tasks within each group. In the 'Merge - visualize' task group (Table~\ref{tab:taskset01}), the easiest and, mostly, solvable, almost all models achieved accuracy over 0.6. The o4-mini had the highest accuracy, followed by Claude Sonnet 3.5, 4, all with success rates over 0.7.
\begin{table}
  \caption{Success rate in solving 'Merge - visualize' tasks}
  (36 tasks from total of 202)
  \label{tab:taskset01}
  \begin{tabular}{lcccc} 
    \toprule
    Model&No&Partial&Match&$\Delta$ in \\
     &match&match& &solution\\
    &&&&length\\  
    \midrule
    Gemini 2.0 Flash&0.32&0.00&0.68&0.2\\
    Gemini 2.5 Pro Preview&0.20&0.13&0.67&0.7\\        
    Claude Haiku 3.5&0.25&0.15&0.60&0.6\\
    Claude Sonnet 3.5&0.17&0.11&0.72&0.3\\  
    Claude Sonnet 4&0.19&0.09&0.71&1.3\\   
    GPT-4o&0.32&0.08&0.60&0.4\\
    GPT-4.1&0.28&0.12&0.60&0.8\\
    o4-mini&0.20&0.03&0.77&0.1\\ 
  \bottomrule
\end{tabular}
\end{table}

In the 'Process - merge - visualize' task group, accuracy starts to decrease while the solution lengths increase (Table~\ref{tab:taskset02}). The top three models by accuracy achieve success rates over 0.6: GPT-4o, Claude Sonnet 3.5, and o4-mini, with o4-mini notably outperforming the other eight models at 0.75 accuracy. The rest have accuracy 0.5 or lower. 

In this task group, the tendency of Claude Sonnet 4 to produce lengthy tasks becomes apparent: its solutions on average are 330\% longer than the reference solutions. Claude Haiku 3.5, the next model by mean difference in length with reference solutions, has the solutions 190\% longer on average. The longer solutions usually mean repeated or failed steps, unnecessary or failed visualizations or unnecessary data loads early in the solution. This is associated with more errors, with the final response sometimes being unusable even if it contains the answer, or with inability to complete solution within the recursion limit. It also confuses evaluators that tend to penalize excessively long solutions despite the instructions not to. Most common problems for these two models were attempts to provide solutions for unsolvable tasks, errors in data filtering, selection of datasets and dates for mapping, lack of awareness of recent updates (like expansion of BRICS) even for models released after the events.

\begin{table}
  \caption{Success rate in solving 'Process - merge - visualize’ tasks}
  (56 tasks from total of 202)
  \label{tab:taskset02}
  \begin{tabular}{lcccc} 
    \toprule
    Model&No&Partial&Match&$\Delta$ in \\
     &match&match& &solution\\
    &&&&length\\  
    \midrule
    Gemini 2.0 Flash&0.46&0.06&0.48&0.04\\
    Gemini 2.5 Pro Preview&0.45&0.05&0.50&1.0\\        
    Claude Haiku 3.5&0.54&0.09&0.36&1.9\\
    Claude Sonnet 3.5&0.36&0.00&0.61&0.3\\  
    Claude Sonnet 4&0.50&0.08&0.41&3.3\\   
    GPT-4o&0.32&0.07&0.61&0.2\\
    GPT-4.1&0.44&0.06&0.51&1.3\\ 
    o4-mini&0.23&0.03&0.75&0.4\\ 
  \bottomrule
\end{tabular}
\end{table}
In the third task group, 'Spatial operations' only o4-mini achieves a success rate 0.56 (Table~\ref{tab:taskset03}). However, in this task group o4-mini rejected 62\% of tasks, which is higher than for the whole set (49\%), hence the shorter mean difference between length of the ground truth and o4-mini solutions. Claude Sonnet 3.5, Gemini 2.5 Pro Preview, GPT-4o follow with the success rates 0.38-0.40. Six of the tested models have solutions on average more than twice as long as the reference solutions and the issues associated with this.

\begin{table}
  \caption{Success rate in solving 'Spatial operations’ tasks}
  (53 tasks from total of 202)
  \label{tab:taskset03}
  \begin{tabular}{lcccc} 
    \toprule
    Model&No&Partial&Match&$\Delta$ in \\
     &match&match& &solution\\
    &&&&length\\  
    \midrule
    Gemini 2.0 Flash&0.65&0.10&0.26&1.9\\
    Gemini 2.5 Pro Preview&0.46&0.16&0.39&0.7\\        
    Claude Haiku 3.5&0.56&0.18&0.26&2.4\\
    Claude Sonnet 3.5&0.50&0.11&0.40&1.6\\  
    Claude Sonnet 4&0.54&0.12&0.34&3.8\\   
    GPT-4o&0.55&0.06&0.38&2.3\\
    GPT-4.1&0.46&0.19&0.35&2.9\\
    o4-mini&0.33&0.10&0.56&-0.08\\ 
  \bottomrule
\end{tabular}
\end{table}
In the fourth task group, 'Spatial Operations', the accuracy (success rate) for most of models is the lowest across all task groups (Table~\ref{tab:taskset04}). The top three models in this task group are: o4-mini with an accuracy rate of 0.60, due to the better ability to identify unsolvable tasks, Gemini 2.5 Pro Preview, and GPT-4o.
\begin{table}
  \caption{Success rate in solving 'Heatmaps, contour lines’ tasks}
  (54 tasks from total of 202)
  \label{tab:taskset04}
  \begin{tabular}{lcccc} 
    \toprule
    Model&No&Partial&Match&$\Delta$ in \\
     &match&match& &solution\\
    &&&&length\\     
    \midrule
    Gemini 2.0 Flash&0.74&0.05&0.21&1.3\\
    Gemini 2.5 Pro Preview&0.60&0.08&0.31&0.8\\        
    Claude Haiku 3.5&0.72&0.13&0.15&2.0\\
    Claude Sonnet 3.5&0.73&0.07&0.20&1.4\\  
    Claude Sonnet 4&0.65&0.15&0.20&4.6\\   
    GPT-4o&0.63&0.07&0.30&1.2\\
    GPT-4.1&0.61&0.13&0.25&2.8\\ 
    o4-mini&0.32&0.09&0.60&0.5\\ 
  \bottomrule
\end{tabular}
\end{table}

\begin{table*}
  \caption{Share of solutions matching across models and tasks groups}
  \label{tab:overall}
  \begin{tabular}{lccccccc}
    \toprule
    Model&Unsolvable&Solvable& Merge - & Process - & Spatial& Heatmaps,\\
    &tasks&tasks&visualize&merge - &operations& contour\\   
    &&&&visualize&lines\\       
    \midrule
    Gemini 2.0 Flash&0.40&0.39&0.68&0.48&0.26&0.21\\
    Gemini 2.5 Pro Preview&0.55&0.39&0.67&0.50&0.39&0.31\\     
    Claude Haiku 3.5&0.22&0.40&0.60&0.36&0.26&0.15\\
    Claude Sonnet 3.5&0.47&0.47&0.72&0.61&0.40&0.20\\  
    Claude Sonnet 4&0.17&0.55&0.71&0.41&0.34&0.20\\    
    GPT-4o&0.62&0.38&0.60&0.61&0.38&0.30\\
    GPT-4.1&0.43&0.42&0.60&0.51&0.35&0.25\\  
    o4-mini&0.90&0.51&0.77&0.75&0.56&0.60\\ 
    \bottomrule
  \end{tabular}
\end{table*}

\subsection{Token usage}
\label{subsec-tokens}
We also compared the number of tokens used by the models to solve the benchmark set (Table~\ref{tab:tokens}) since it influences costs and costs are a critical parameter in practice. Note that because solving the tasks is an iterative process and at every iteration the previous model responses are fed back into model, the number of generated output tokens correlates with measured input tokens.
What is clear is that an increase in token usage and generation does not necessarily translate into accuracy in solving tasks. 

Anthropic models were on the higher end of the input/output tokens used. Claude Sonnet 4, due to its reasoning ability, creates 2-3 times more input tokens and up to 5 times more output tokens than competitors, yet achieves similar or worse accuracy than models like Claude Sonnet 3.5, GPT-4o, GPT-4.1 and Gemini 2.5 Pro Preview (Table~\ref{tab:overall}). Anthropic models are more expensive in use: the most affordable of the three, Claude 3.5 Haiku, due to the amount of tokens generated ended up costing more than any OpenAI or Google models tested in this experiment to solve the benchmark set.

Notably, the best performed on this benchmark set OpenAI's o4-mini model was among the three models with the lowest amount of input tokens used and the highest amount output token (3 times of the Claude Sonnet 4 output and 8 times GPT-41, next OpenAI's model by output).

% A final note on the costs of solving the benchmark set, as this is an important consideration for developing geospatial agents  (Figure~\ref{fig:costs}). At the pricing in effect at the time of the experiment, the best models to use in geospatial agents based on cost-success rate trade-offs are OpenAI's o4-mini and Google's Gemini 2.0 Flash: the former for high success rates with moderate costs, and the latter for extremely low costs with a comparatively decent success rate (Figure~\ref{fig:costs}, Table~\ref{tab:overall}). Anthropic models are the most expensive, with Claude Sonnet 4 costs far exceeding those of any other model; therefore, even with high success rates, careful consideration should be given to efficient ways to use them in geospatial agents. Costs of running the remaining models on the benchmark set were similar to each other.

\begin{table}
  \caption{Tokens used for generation of solutions for the whole benchmark set (single run)}
  \label{tab:tokens}
  \begin{tabular}{lcc} 
    \toprule
    Model&Input&Output\\
     &tokens (mln)&tokens (k) \\
    \midrule
    Gemini 2.0 Flash&7.5&40.7\\
    Gemini 2.5 Pro Preview&10.4&85.4\\     
    Claude Haiku 3.5&14.9&196.0\\
    Claude Sonnet 3.5&12.2&151.2\\  
    Claude Sonnet 4&26.9&291.6\\    
    GPT-4o&8.0&61.2\\
    GPT-4.1&12.7&121.3\\ 
    o4-mini&7.9&956.6\\ 
  \bottomrule
\end{tabular}
\end{table}

% \begin{figure}
%     \centering
%     \includegraphics[width=1\linewidth]{generation-tokens.png}
%     \caption{Token usage for solving tasks across benchmarked LLMs (single run)}
%     \label{fig:token-usage}
% \end{figure}

% \begin{figure}
%     \centering
%     \includegraphics[width=1\linewidth]{generation-costs.png}
%     \caption{Costs for solving the benchmark set by LLM (single run)}
%     \label{fig:costs}
% \end{figure}

\subsection{Error analysis}
Looking at cases where candidate solutions did not match reference, the main causes/types of errors were: 
\begin{enumerate}[leftmargin=1em]
    \item Poor understanding of geometrical relations: e.g., taking values from centroid point of a polygon instead of calculating average value over polygon, trying to use centroids of lines while looking for geometries overlaps, not understanding that object's centroid can lie outside the boundary of the object, etc.
    \item Lack of recent world knowledge: LLMs worked well with established international groupings, but struggled with ambiguous groupings (Upper/Lower Nile countries) or with more recent changes (e.g., BRICS expansion in 2024), even if the event happened before the model's cut-off date.
    \item Over-reliance on the world knowledge: using internally known spelling of names rather than dataset values for filtering, despite having tools to check column values. 
    \item Inefficient data manipulations: merging without prefiltering (e.g., merging with all USA counties instead of first filtering to specific state); filtering by countries when filtering by continent/region was possible, and/or repeating filtering steps sometimes confusing LLM and derailing solutions.
    \item Skipping steps that would narrow down the geodatasets to the required geography. E.g keeping all the North American lakes when the question was about USA.
    \item Attempting to solve with partial data, trying to be helpful by using whatever geography was available instead of rejecting tasks with insufficient data coverage.
    \item Modifying tasks to make them solvable rather than solving the original problem.
    \item Hallucinating unjustified data processing steps, rejecting solvable tasks in the middle of solution, using some numbers for thresholds or names of countries in filters without justification or connection with provided datasets.
    \item Answering/rejecting the task without using the tools and dataset provided.
\end{enumerate}
 Looking at models' mistakes, it is likely that all models will perform much better if there is human-in-the-loop to disambiguate vague situations or provide step-by-step instructions on how to solve the task and more detailed explanations of the geographical scope of the tasks. It was also observed that making the tools more narrow specialized improved the accuracy of the agent.  

\section{Conclusions}
We have created a benchmark dataset for multistep geospatial tasks, introduced an evaluation framework and evaluated the leading mainstream commercial LLMs on these tasks.

The analysis indicates that the commercially available LLMs with function calling can be used successfully to solve geospatial tasks even when tasks are stated casually, with ambiguity, missing information, and minimal initial instructions. The performance of the models varies by the complexity of geospatial tasks, the degree of spatial thinking required to solve them, and how strongly the model is skewed to provide any answer at all rather than reject an unsolvable task.

In terms of comparative performance (see Table~\ref{tab:overall} and Figure~\ref{fig:summary-matches}), the OpenAI's o4-mini showed the best results overall. Anthropic's Claude Sonnet 3.5 ranked second overall by consistently placing in the top three for accuracy across multiple categories. 

The three other models with top-3 accuracy in 2-3 categories showed clear trade-off between the ability to solve tasks and to identify unsolvable ones: Claude Sonnet 4 performs more than three times better on solvable tasks than on unsolvable ones thus losing in accuracy in separate task groups; GPT-4o and Gemini 2.5 Pro are better at correctly rejecting tasks than at solving them and thus have better accuracy on task groups. 

The remaining models, GPT-4.1 and Google's Gemini 2.0 Flash, are more balanced in rejecting and solving tasks and have lower accuracy across task groups. 

Based on the cost of running the models, the best options for geospatial agents might be OpenAI's o4-mini due to its combination of moderate costs and high accuracy, and Google's Gemini 2.0 Flash as the most cost-efficient model while still providing decent results. Anthropic models performed well on easy tasks and solvable tasks in general; however, their preference for trying to provide any solution rather than rejecting, high token usage, and price might be disadvantages when using them for geospatial tasks. 

\section{Limitations}
We did not tune prompts individually for each LLM; we did not provide human feedback to the model to guide solution; we evaluate only solution algorithm not the final response (for rejects and final numbers); the benchmark set size is moderate; we did not evaluate the variance of the responses (only the binomial confidence bounds).

% \begin{acks}
% The authors thank their families for their patience and support.
% \end{acks}

\bibliographystyle{ACM-Reference-Format}
\bibliography{GeoBenchX}

\appendix

\section*{Appendix A: Prompts for solver agent}
\label{appendix-prompts}

System prompt is in Figure \ref{fig:system-prompt}, rules prompt is in Figure \ref{fig:rules-prompt}.

% Set up listings for line wrapping
\lstset{
  breaklines=true,
  breakatwhitespace=false,
  basicstyle=\small\ttfamily,
  columns=flexible,
  keepspaces=true,
  xleftmargin=0pt,
  framexleftmargin=0pt,
  framextopmargin=0pt,
  frame=none,
  framesep=0pt,
  breakindent=0pt,
  postbreak={}
}

\begin{figure*}[!ht]
    \centering
    \begin{tcolorbox}[
        width=\linewidth,
        colback=gray!5,
        colframe=gray!40,
        title=System Prompt,
        fonttitle=\bfseries
    ]
    \begin{lstlisting}[backgroundcolor=\color{gray!5}, aboveskip=0pt, belowskip=0pt]
You are a geographer who answers the questions with maps whenever possible. You do have a list of python functions that help you to load statistical data, geospatial data, rasters, merge dataframes, filter dataframes, get uniques values from columns, to plot a map, make spatial operations on raster and vector data. You do not provide code to generate a map, you provide names of the datasets you used; the statistical dataset, the geospatial dataset and the resulting map with legend and needed explanations. Today is {current_date}.
    \end{lstlisting}
    \end{tcolorbox}
    \caption{System prompt for solver agent.}
    \label{fig:system-prompt}
\end{figure*}

\begin{figure*}[!ht]
    \centering
    \begin{tcolorbox}[
        width=\linewidth,
        colback=gray!5,
        colframe=gray!40,
        title=Rules Prompt,
        fonttitle=\bfseries
    ]
    \begin{lstlisting}[backgroundcolor=\color{gray!5}]
While solving the tasks, please, follow the next rules:
- If a task is not a geospatial task, if it does not require calling provided tools and use provided datasets to solve, proceed with reply, no explanations needed, no need to call 'reject_task'.
- If the task is a geospatial task, however, this task can NOT be solved with either available datasets or tools, call the tool 'reject_task'.
- If user did not specify the date of the data to be plotted on the map, you map the latest available period that has non empty data for more than 70% of the objects in the datasets (example: if a question is about countries of Africa, you map the latest year that has data for more than 70% of countries in Africa).
- For bivariate maps select the data for the same year for both variables. For instance, if for variable 1 the latest data are for 2023 and for variable 2 the latest data are for 2014, select data for 2014 for both variable 1 and variable 2. If the common year for which both datasets have data is 2012, take the data for 2012.
- If available dataset or datasets cover only part of the requested geography, consider the data unavailable.
- Do not ask questions to the user, proceed with solving the task till a result is achieved. When the information is missing try to infer the reasonable values.
    \end{lstlisting}
    \end{tcolorbox}
    \caption{Rules prompt for solver agent.}
    \label{fig:rules-prompt}
\end{figure*}

\section*{Appendix B: GeoSpatial Tools List}
\begin{description}
\item[load_data] Loads statistical data from the catalog into a Pandas DataFrame.
\begin{itemize}
\item dataset: Name of the dataset to load
\item output_dataframe_name: Name for the new DataFrame
\item state: Current graph state containing (Geo)DataFrames and messages
\end{itemize}
\item[load_geodata] Loads vector geospatial data from the catalog into a GeoPandas GeoDataFrame.
\begin{itemize}
\item geodataset: Name of the geodataset to load
\item output_geodataframe_name: Name for the new GeoDataFrame
\item state: Current graph state containing (Geo)DataFrames and messages
\end{itemize}
\item[get_raster_path] Constructs path to raster data from the catalog.
\begin{itemize}
\item rasterdataset: Name of the raster dataset to access
\item state: Current graph state containing (Geo)DataFrames and messages
\end{itemize}
\item[get_raster_description] Gets description of a raster dataset including metadata and statistics.
\begin{itemize}
\item raster_path: Path to the raster file
\item output_variable_name: Optional name for storing results
\item state: Current graph state containing (Geo)DataFrames and messages
\end{itemize}
\item[analyze_raster_overlap] Analyzes overlap between two rasters and calculates statistics.
\begin{itemize}
\item raster1_path: Path to first raster file
\item raster2_path: Path to second raster file
\item output_variable_name: Name for storing results
\item resampling_method1/2: Resampling methods
\item plot_result: Whether to display visualization
\item state: Current graph state containing (Geo)DataFrames and messages
\end{itemize}
\item[get_values_from_raster_with_geometries] Masks a raster using vector geometries and calculates statistics.
\begin{itemize}
\item raster_path: Path to the raster file
\item geodataframe_name: Name of GeoDataFrame with geometries
\item output_variable_name: Name for storing results
\item plot_result: Whether to display visualization
\item state: Current graph state containing (Geo)DataFrames and messages
\end{itemize}
\item[merge_dataframes] Merges statistical and geospatial DataFrames using key columns.
\begin{itemize}
\item dataframe_name: Name of DataFrame with statistical data
\item geodataframe_name: Name of GeoDataFrame with spatial data
\item statkey: Column name in statistical data for joining
\item geokey: Column name in spatial data for joining
\item output_dataframe_name: Name for the merged DataFrame
\item state: Current graph state containing (Geo)DataFrames and messages
\end{itemize}
\item[get_unique_values] Gets unique values from a column in a DataFrame/GeoDataFrame.
\begin{itemize}
\item dataframe_name: Name of DataFrame to analyze
\item column: Column name to get unique values from
\item state: Current graph state containing (Geo)DataFrames and messages
\end{itemize}
\item[filter_categorical] Filters DataFrame/GeoDataFrame by categorical values.
\begin{itemize}
\item dataframe_name: Name of DataFrame to filter
\item filters: Dictionary of column names and values
\item output_dataframe_name: Name for filtered DataFrame
\item state: Current graph state containing (Geo)DataFrames and messages
\end{itemize}
\item[filter_numerical] Filters DataFrame/GeoDataFrame using numerical conditions.
\begin{itemize}
\item dataframe_name: Name of DataFrame to filter
\item conditions: Query string with numerical conditions
\item output_dataframe_name: Name for filtered DataFrame
\end{itemize}
\item[calculate_column_statistics] Calculates summary statistics for a numerical column.
\begin{itemize}
\item dataframe_name: Name of DataFrame to analyze
\item column_name: Column to analyze
\item output_variable_name: Name for storing results
\item include_quantiles: Whether to include standard quantiles
\item additional_quantiles: Additional quantiles to calculate
\item state: Current graph state containing (Geo)DataFrames and messages
\end{itemize}
\item[create_buffer] Creates buffer zones around geometries in a GeoDataFrame.
\begin{itemize}
\item geodataframe_name: Name of source GeoDataFrame
\item buffer_size: Size of the buffer in meters
\item output_geodataframe_name: Name for result GeoDataFrame
\item state: Current graph state containing (Geo)DataFrames and messages
\end{itemize}
\item[make_choropleth_map] Creates a choropleth map visualization from a GeoDataFrame.
\begin{itemize}
\item dataframe_name: Name of GeoDataFrame with map data
\item mappingkey: Column to visualize
\item legendtext: Legend title
\item colormap: Color scheme for the map
\item state: Current graph state containing (Geo)DataFrames and messages
\end{itemize}
\item[filter_points_by_raster_values] Samples raster values at point locations and filters by threshold.
\begin{itemize}
\item raster_path: Path to the raster file
\item points_geodataframe_name: Name of GeoDataFrame with points
\item value_column: Column to store raster values
\item output_geodataframe_name: Name for filtered GeoDataFrame
\item filter_type: Type of threshold filter
\item threshold1: First threshold value
\item threshold2: Optional second threshold value
\item plot_result: Whether to display visualization
\item state: Current graph state containing (Geo)DataFrames and messages
\end{itemize}
\item[select_features_by_spatial_relationship] Selects features based on spatial relationships.
\begin{itemize}
\item features_geodataframe_name: GeoDataFrame to select from
\item reference_geodataframe_name: GeoDataFrame to select by
\item spatial_predicates: List of spatial relationships
\item output_geodataframe_name: Name for result GeoDataFrame
\item plot_result: Whether to display visualization
\item state: Current graph state containing (Geo)DataFrames and messages
\end{itemize}
\item[calculate_line_lengths] Calculates lengths of line features in kilometers.
\begin{itemize}
\item geodataframe_name: Name of GeoDataFrame with line features
\item output_variable_name: Name for storing results
\item state: Current graph state containing (Geo)DataFrames and messages
\end{itemize}
\item[calculate_columns] Performs operations between columns of DataFrames.
\begin{itemize}
\item dataframe1_name: Name of first DataFrame
\item column1_name: Column from first DataFrame
\item dataframe2_name: Name of second DataFrame
\item column2_name: Column from second DataFrame
\item operation: Mathematical operation to perform
\item output_column_name: Name for results column
\item output_dataframe_name: Name for result DataFrame
\item state: Current graph state containing (Geo)DataFrames and messages
\end{itemize}
\item[scale_column_by_value] Performs operations between a column and a value.
\begin{itemize}
\item dataframe_name: Name of DataFrame to modify
\item column_name: Column to scale
\item operation: Mathematical operation
\item value: Numerical value for scaling
\item output_column_name: Name for results column
\item output_dataframe_name: Name for result DataFrame
\item state: Current graph state containing (Geo)DataFrames and messages
\end{itemize}
\item[make_heatmap] Creates an interactive heatmap from point data.
\begin{itemize}
\item geodataframe_name: Name of GeoDataFrame with points
\item value_column: Column for intensity values
\item center_lat/lon: Map center coordinates
\item zoom_level: Initial zoom level
\item radius: Radius of influence for each point
\item map_style: Base map style
\item width/height: Map dimensions
\item state: Current graph state containing (Geo)DataFrames and messages
\end{itemize}
\item[visualize_geographies] Displays multiple GeoDataFrames as map layers.
\begin{itemize}
\item geodataframe_names: List of GeoDataFrame names
\item layer_styles: Optional styling for each layer
\item basemap_style: Style of the basemap
\item title: Map title
\item add_legend: Whether to add a legend
\item state: Current graph state containing (Geo)DataFrames and messages
\end{itemize}
\item[get_centroids] Calculates centroids of polygon features.
\begin{itemize}
\item geodataframe_name: Name of GeoDataFrame with polygons
\item output_geodataframe_name: Name for centroid GeoDataFrame
\item state: Current graph state containing (Geo)DataFrames and messages
\end{itemize}
\item[generate_contours_display] Creates contour lines using GDAL algorithm.
\begin{itemize}
\item raster_path: Path to the raster file
\item output_filename: Name for output shapefile
\item contour_interval: Interval between contour lines
\item column_title: Name for values column
\item nodataval: Value for no-data pixels
\item output_geodataframe_name: Name for contour GeoDataFrame
\item min/max_value: Range for contours
\item plot_result: Whether to display visualization
\item plot_title: Title for the plot
\item cleanup_files: Whether to remove temporary files after processing
\item colormap: Matplotlib colormap name
\item figsize: Figure size as JSON array [width, height]
\item add_colorbar: Whether to add a colorbar to the plot
\item plot_background: Whether to display raster as background
\item state: Current graph state containing (Geo)DataFrames and messages
\end{itemize}
\item[make_bivariate_map] Creates a bivariate choropleth map showing relationship between two variables.
\begin{itemize}
\item dataframe_name: Name of GeoDataFrame with map data
\item var1/var2: Variable column names
\item var1_name/var2_name: Display names for variables
\item title: Map title
\item output_variable_name: Name for storing results
\item state: Current graph state containing (Geo)DataFrames and messages
\end{itemize}
\item[reject_task] Returns a standardized message when task cannot be solved.
\begin{itemize}
    \item state: Current graph state containing (Geo)DataFrames and messages
\end{itemize}
\end{description}

\end{document}